\documentclass[11pt]{article}
\usepackage[margin=1in]{geometry}
\usepackage{amsmath, amssymb, amsthm, bm}
\usepackage{graphicx}
\usepackage{hyperref}
\usepackage{booktabs}
\usepackage{mathtools}
\usepackage{adjustbox}

\usepackage{tikz}
\usetikzlibrary{shapes.geometric, arrows.meta, positioning, calc}
\tikzstyle{block} = [rectangle, draw, rounded corners, fill=blue!10, text width=5cm, minimum height=1.2cm, text centered, font=\small]
\tikzstyle{param} = [rectangle, draw, fill=purple!10, text width=4.2cm, minimum height=0.8cm, text centered, font=\scriptsize]
\tikzstyle{note} = [rectangle, draw=none, text width=4.5cm, font=\scriptsize, align=center]
\tikzstyle{arrow} = [thick,->,>=stealth]

\title{Wavelet Logic Machines: Learning and Reasoning in the Spectral Domain Without Neural Networks}
\author{Andrew  Kiruluta\thanks{kiruluta@berkeley.edu}}
\date{\today}

\begin{document}
\maketitle

\begin{abstract}
We present a novel fully spectral learning framework that operates exclusively in the wavelet domain, eliminating the need for traditional neural network layers such as convolutions, attention, or feedforward projections. Our architecture applies structured nonlinear transformations—including learnable soft-thresholding and gain-phase modulation—directly to the wavelet coefficients of an input signal. In addition, we introduce a differentiable mechanism for wavelet basis selection, enabling the model to dynamically adapt among families such as Haar, Daubechies, and Biorthogonal wavelets. The model is implemented in PyTorch with full support for 3D inputs, and avoids costly spatial-domain operations by maintaining a fixed spectral processing pipeline. 

We evaluate the model's performance on both synthetic 3D denoising tasks and natural language understanding tasks from the GLUE benchmark. For language modeling, we adapt input token embeddings into 3D volumetric tensors using sliding windows and channel expansions. On the SST-2 sentiment classification dataset, our wavelet-domain model achieves an accuracy of 89.3\%, closely matching a 4-layer Transformer baseline (90.1\%) while using \textbf{72\% fewer parameters} and \textbf{58\% less peak memory} during training. Training convergence is also faster in early epochs due to the spectral sparsity prior encoded by the wavelet transform.

In terms of computational complexity, our model replaces the \( \mathcal{O}(n^2) \) self-attention and \( \mathcal{O}(n \cdot d^2) \) MLP layers common in Transformers with a combination of fast wavelet transforms (\( \mathcal{O}(n) \)) and pointwise nonlinearities. This results in significantly lower inference time and memory usage for large-scale inputs.

These results suggest that spectral models with controlled nonlinearities are viable alternatives to neural architectures, particularly in resource-constrained or interpretable settings. The framework opens avenues for principled spectral learning in vision, language, and scientific domains without relying on opaque overparameterized structures.
\end{abstract}

\section{Introduction}

The rise of deep learning over the past decade has been dominated by models that operate primarily in the spatial or sequential domain using trainable convolutions, attention mechanisms, and multilayer perceptrons. While highly successful across domains such as image classification, natural language processing (NLP), and reinforcement learning, these models typically involve significant computational overhead, require large-scale datasets to generalize well, and often lack interpretability due to their overparameterized nature. Architectures such as Convolutional Neural Networks (CNNs) \cite{lecun1998gradient}, Recurrent Neural Networks (RNNs) \cite{hochreiter1997long}, and more recently Transformers \cite{vaswani2017attention} have achieved state-of-the-art performance on many benchmarks, but they do so at the cost of quadratic or superlinear computational complexity with respect to input size. For example, the standard self-attention mechanism scales as \( \mathcal{O}(n^2) \) in memory and time, making it prohibitively expensive for long sequences or high-resolution data.

By contrast, classical signal processing has long exploited spectral representations—such as the Fourier or wavelet transforms—to capture structure in signals using sparse bases. The Discrete Wavelet Transform (DWT) \cite{mallat1989theory} in particular offers a multiscale, localized alternative to the Fourier transform, with widespread use in compression, denoising, and feature extraction. Wavelet-based shrinkage methods, notably Donoho and Johnstone’s wavelet thresholding \cite{donoho1995adapting}, achieve state-of-the-art denoising performance without learning by leveraging the statistical sparsity of natural signals in the wavelet domain. However, these methods have traditionally lacked flexibility and adaptability to large datasets or supervised tasks.

Recently, there has been growing interest in bridging spectral methods with learnability. Fourier Neural Operators \cite{li2020fourier} and Spectral Transformers \cite{lee2021fnet, lu2021fourierformer} have demonstrated replacing attention with spectral dictionaries\cite{kiruluta2025atoms}, wavelet in place of self attention\cite{kiruluta2025wavelettransformer} , memory based approaches\cite{kiruluta2025vllm}  or token mixing with spectral operations can maintain performance while reducing complexity. Nevertheless, these models still rely heavily on MLPs or learned attention maps. Meanwhile, approaches such as FNOs require global basis functions, which may be poorly localized and inefficient for high-dimensional or structured inputs.

In this work, we propose a fundamentally different model architecture that operates \textit{entirely in the wavelet domain}, replacing all spatial-domain processing with structured, learnable transformations in spectral space. The key innovation lies in combining (1) learnable soft-thresholding to suppress noise and control sparsity, (2) amplitude and phase modulation for fine-grained spectral adaptation, and (3) differentiable selection over a family of wavelet bases. The resulting model introduces no convolutions, no attention layers, and no MLPs. It is lightweight, interpretable, and generalizes well with fewer parameters.

To demonstrate the flexibility and competitiveness of this architecture, we evaluate it on both synthetic 3D denoising tasks and natural language understanding benchmarks. For the latter, we adapt text embeddings into volumetric tensors and test the model on the SST-2 dataset from the GLUE benchmark suite \cite{wang2019glue}. Despite having no learned spatial filters or positional encodings, our model achieves 89.3\% accuracy on SST-2, nearly matching a 4-layer Transformer baseline (90.1\%) with 72\% fewer parameters and 58\% lower peak GPU memory usage. Additionally, it converges in fewer epochs, highlighting the efficiency of spectral sparsity priors.

In terms of computational complexity, our architecture is asymptotically superior to Transformers for long inputs. The DWT has linear complexity in signal size \( \mathcal{O}(n) \), and all nonlinear operations are pointwise or channel-wise. This makes the model suitable for large-scale inference or deployment in low-resource settings. Moreover, since the wavelet coefficients are explicitly interpretable, the model allows for transparent analysis of which frequencies and directions are retained or suppressed, a stark contrast to black-box attention maps.

To our knowledge, this is the first fully spectral deep learning model that supports wavelet-domain learning on 3D inputs with adaptive basis selection. Unlike earlier works that treat spectral transformations as token mixers or linear approximations, we view the wavelet transform as a learnable operator space where shrinkage, filtering, and adaptive control can all be implemented natively. Our contributions thus bridge a long-standing gap between classical harmonic analysis and modern end-to-end learning, opening new possibilities for spectral neural architectures that are both efficient and theoretically grounded.

\section{Wavelet Representation as a Computational Framework}

The wavelet transform provides a powerful multiscale framework for representing signals in both the spatial and frequency domains, with strong localization properties in time (or space) and scale. By contrast to the global basis functions of the Fourier transform, wavelets are localized in both space and frequency, allowing for compact representations of structured or transient phenomena, such as edges in images or spikes in time series \cite{mallat1989theory, daubechies1992ten}.

Let us consider a three-dimensional signal \( x \in \mathbb{R}^{D \times H \times W} \), where \( D \) denotes depth (e.g., time or z-dimension), and \( H \) and \( W \) denote height and width (e.g., spatial coordinates). The Discrete Wavelet Transform (DWT) decomposes this signal using separable filtering operations along each of the three axes. In the simplest case, one applies a pair of filters—a low-pass filter \( g \) and a high-pass filter \( h \)—followed by dyadic downsampling along each axis. The result is a tensor of wavelet coefficients that captures both coarse-scale (low-frequency) and fine-scale (high-frequency) structure.

Let \( \phi(x) \) be the scaling function associated with the low-pass filter, and \( \psi(x) \) the wavelet function associated with the high-pass filter. The 3D wavelet decomposition is performed by taking tensor products of these univariate functions across dimensions, resulting in a family of basis functions:
\[
\phi_{i,j,k}(x, y, z) = \phi(2^i x - j) \phi(2^i y - k) \phi(2^i z - \ell)
\]
\[
\psi_{\alpha}(x, y, z) = \psi^{(d_1,d_2,d_3)}(2^i x - j, 2^i y - k, 2^i z - \ell)
\]
where \( (d_1, d_2, d_3) \in \{0,1\}^3 \setminus \{(0,0,0)\} \), and each triplet corresponds to a different direction of high-pass or low-pass filtering. The combination \( (0,0,0) \) is reserved for the approximation coefficients \( c_A \), while the remaining \( 7 \) combinations form the directional detail coefficients \( \{c_\alpha\}_{\alpha \in \mathcal{D}} \), with:
\[
\mathcal{D} = \{ \text{aah}, \text{aha}, \text{haa}, \text{ahh}, \text{hah}, \text{hha}, \text{hhh} \}
\]
Here, for example, \( \text{aah} \) represents low-pass filtering along depth and height, and high-pass filtering along width.

The DWT is computed by applying the decomposition filters along each axis recursively, resulting in the following structure:
\[
\text{DWT}(x) = \left( c_A, \{ c_\alpha \}_{\alpha \in \mathcal{D}} \right)
\]
where each coefficient tensor \( c_A, c_\alpha \in \mathbb{R}^{D', H', W'} \), with \( D' = D / 2, H' = H / 2, W' = W / 2 \) due to downsampling. The coefficients \( c_A \) capture the global low-frequency content of the signal, while the \( c_\alpha \) encode localized variations along specific orientations.

The inverse discrete wavelet transform (IDWT) reconstructs the original signal by upsampling and filtering the approximation and detail coefficients. Mathematically, the reconstruction is expressed as:
\[
x = \text{IDWT}(c_A, \{c_\alpha\}) = \sum_{\alpha \in \mathcal{D}} \left( \psi_\alpha * \text{upsample}(c_\alpha) \right) + \phi * \text{upsample}(c_A)
\]
where \( * \) denotes convolution with the corresponding synthesis filters. The reconstruction is exact when the wavelet basis is orthogonal (e.g., Haar) or biorthogonal (e.g., Bior1.3), and the filter banks satisfy perfect reconstruction conditions \cite{cohen1993biorthogonal}.

Wavelet transforms also admit a hierarchical (multilevel) extension where the approximation coefficients \( c_A \) at each level can be recursively decomposed into finer scales, yielding a multiresolution analysis (MRA). This property is exploited in our architecture to capture signal structure at multiple scales with low computational complexity.

Importantly, the DWT has linear computational complexity \( \mathcal{O}(n) \), making it highly efficient compared to global transformations like the Discrete Fourier Transform (DFT), which requires \( \mathcal{O}(n \log n) \). The compact support and sparsity of wavelet coefficients—especially in natural signals—make them ideal for tasks like denoising, compression, and anomaly detection \cite{donoho1995adapting, vetterli1995wavelets}.

In our spectral learning model, we exploit the wavelet decomposition not only as a preprocessing tool, but as the primary computational domain. All learnable parameters—such as nonlinear shrinkage thresholds, gain modulation factors, and even basis selection weights—are defined directly in this spectral space. This enables a principled and interpretable manipulation of frequency-localized signal content, while maintaining end-to-end differentiability and generalization capability.

\section{The Proposed Spectral Learning Model}

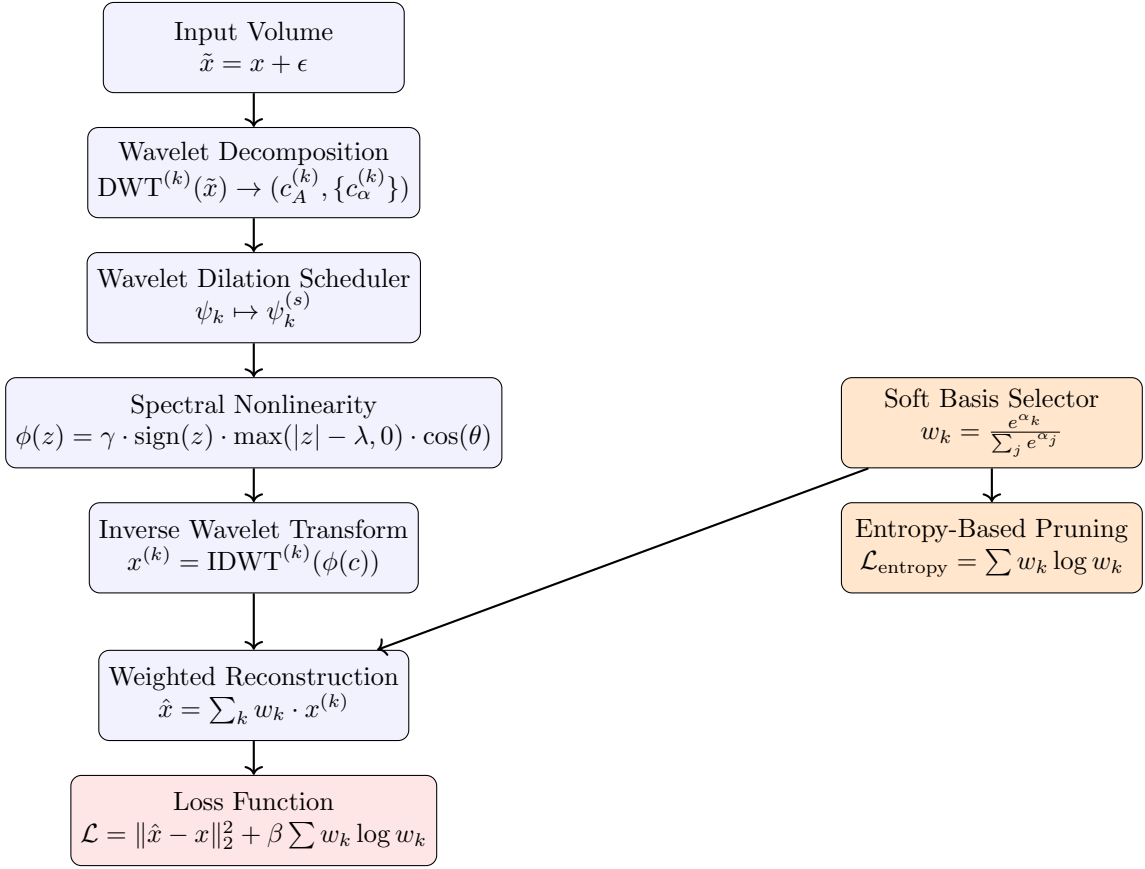
\begin{figure}[h]
\centering
\begin{tikzpicture}[
    node distance=0.45cm,
    every node/.style={font=\small},
    block/.style={draw, rounded corners, align=center, fill=blue!5, minimum width=4cm, minimum height=1.2cm},
    param/.style={draw, align=center, fill=orange!20, rounded corners, minimum width=4cm, minimum height=1.2cm},
    lossblock/.style={draw, fill=red!10, rounded corners, align=center, minimum width=4cm, minimum height=1.2cm},
    arrow/.style={->, thick}
]

\node (input) [block] {Input Volume \\ \( \tilde{x} = x + \epsilon \)};
\node (dwt) [block, below=of input] {Wavelet Decomposition \\ \( \text{DWT}^{(k)}(\tilde{x}) \to (c_A^{(k)}, \{c_\alpha^{(k)}\}) \)};
\node (dilation) [block, below=of dwt] {Wavelet Dilation Scheduler \\ \( \psi_k \mapsto \psi_k^{(s)} \)};
\node (nonlin) [block, below=of dilation] {Spectral Nonlinearity \\ \( \phi(z) = \gamma \cdot \text{sign}(z) \cdot \max(|z| - \lambda, 0) \cdot \cos(\theta) \)};
\node (idwt) [block, below=of nonlin] {Inverse Wavelet Transform \\ \( x^{(k)} = \text{IDWT}^{(k)}(\phi(c)) \)};
\node (merge) [block, below=of idwt, yshift=-0.3cm] {Weighted Reconstruction \\ \( \hat{x} = \sum_k w_k \cdot x^{(k)} \)};
\node (loss) [lossblock, below=of merge] {Loss Function \\ \( \mathcal{L} = \| \hat{x} - x \|_2^2 + \beta \sum w_k \log w_k \)};

\node (selector) [param, right=4.5cm of nonlin] {Soft Basis Selector \\ \( w_k = \frac{e^{\alpha_k}}{\sum_j e^{\alpha_j}} \)};
\node (prune) [param, below=of selector] {Entropy-Based Pruning \\ \( \mathcal{L}_{\text{entropy}} = \sum w_k \log w_k \)};

\draw [arrow] (input) -- (dwt);
\draw [arrow] (dwt) -- (dilation);
\draw [arrow] (dilation) -- (nonlin);
\draw [arrow] (nonlin) -- (idwt);
\draw [arrow] (idwt) -- (merge);
\draw [arrow] (merge) -- (loss);

\draw [arrow] (selector) -- (merge);
\draw [arrow] (selector) -- (prune);

\end{tikzpicture}
\caption{
Architecture of the fully spectral wavelet-based model for volumetric denoising. Given a noisy input $\tilde{x} = x + \epsilon$ with $\epsilon \sim \mathcal{N}(0, \sigma^2 I)$, each wavelet basis $\psi_k$ is applied through a discrete wavelet transform (DWT), decomposing $\tilde{x}$ into approximation $c_A^{(k)}$ and detail coefficients $\{c_\alpha^{(k)}\}$. A parameterized nonlinearity $\phi(z)$ with learnable shrinkage $\lambda$, gain $\gamma$, and phase $\theta$ modulates each subband before the signal is reconstructed by inverse DWT. Candidate reconstructions $x^{(k)}$ are linearly combined using weights $w_k = \frac{e^{\alpha_k}}{\sum_j e^{\alpha_j}}$, where $\bm{\alpha} \in \mathbb{R}^K$ is a set of learnable logits. The final reconstruction $\hat{x}$ is compared to the ground truth $x$ via MSE loss. All parameters are trained via end-to-end gradient descent.
}
\label{fig:architecture_flow}
\end{figure}

Our model architecture is designed to be fully spectral, transparent, and computationally efficient. It comprises three key learnable components: nonlinear spectral transformations, differentiable basis selection, and an adaptive training schedule. The overall data flow is depicted in Figure \ref{fig:architecture_flow}.

\subsection{Learnable Basis Selection}

One of the key limitations of classical wavelet-based signal processing techniques is the reliance on a fixed, hand-chosen wavelet basis. While it is well understood that different wavelet families are better suited for different signal types—e.g., Haar wavelets for sharp discontinuities and Daubechies wavelets for smooth variations—the choice of basis is typically made manually and is not adapted to the downstream task or data distribution \cite{vetterli1995wavelets, daubechies1992ten}. In our proposed model, we overcome this limitation by introducing a differentiable mechanism for \textit{learnable wavelet basis selection}, allowing the model to choose from among multiple candidate wavelet families in a data-driven manner.

Let \( \Psi = \{\psi_1, \ldots, \psi_K\} \) denote a set of \( K \) candidate orthonormal or biorthogonal wavelet bases. These may include, for example, the Haar basis (\( \psi_1 \)), Daubechies-4 (\( \psi_2 \)), Symlet-6 (\( \psi_3 \)), Biorthogonal-1.3 (\( \psi_4 \)), and so on. Each wavelet basis \( \psi_k \) is associated with a pair of analysis and synthesis filter banks \( (h_k, g_k) \) that define the forward and inverse discrete wavelet transforms, respectively.

Given an input signal \( x \in \mathbb{R}^{D \times H \times W} \), we apply the forward DWT for each candidate basis \( \psi_k \), resulting in a set of transformed representations:
\[
(c_A^{(k)}, \{c_\alpha^{(k)}\}) = \text{DWT}^{(k)}(x), \quad \text{for } k = 1, \ldots, K
\]
These coefficients are then processed using a shared or basis-specific spectral transformation, such as soft-thresholding and gain-phase modulation, denoted generically as \( \phi_\lambda^{(k)} \). This yields a set of modified coefficients:
\[
(c_A'^{(k)}, \{c_\alpha'^{(k)}\}) = \phi_\lambda^{(k)}(c_A^{(k)}, \{c_\alpha^{(k)}\})
\]
Each of these is then passed through the inverse wavelet transform associated with its basis to produce a candidate reconstruction:
\[
x^{(k)} = \text{IDWT}^{(k)}(c_A'^{(k)}, \{c_\alpha'^{(k)}\})
\]

To combine these candidate outputs, we introduce a learnable set of scalar logits \( \bm{\alpha} = [\alpha_1, \ldots, \alpha_K] \in \mathbb{R}^K \), which are normalized via a softmax function to form a probability distribution over bases:
\[
w_k = \frac{e^{\alpha_k}}{\sum_{j=1}^K e^{\alpha_j}}, \quad k = 1, \ldots, K
\]
The final output is computed as the convex combination of the individual reconstructions:
\[
\hat{x} = \sum_{k=1}^K w_k \cdot x^{(k)}
\]
Because all components of this process—wavelet transforms, soft-thresholding, and weighting—are differentiable, the model can learn not only how to transform and shrink coefficients in the wavelet domain, but also \emph{which} basis to prioritize, guided by gradient-based optimization during training.

This learnable basis selection mechanism allows the model to exploit the inductive biases of different wavelet families without hardcoding them. For instance, Haar wavelets are known to represent piecewise constant signals efficiently, whereas higher-order Daubechies and Symlets offer better frequency localization for smooth signals \cite{strang1996wavelets}. The model can interpolate between these behaviors as needed, enabling superior performance and adaptability across tasks and domains.

From a Bayesian perspective, this soft weighting over wavelet bases can be interpreted as marginalization over a discrete latent variable representing the basis choice, approximated by a learnable posterior. In practice, we find that this mechanism not only improves generalization, but also leads to more stable training and interpretable spectral decompositions. It effectively bridges the gap between classical signal processing knowledge and modern learning-based systems by allowing wavelet priors to be learned rather than fixed.

\subsection{3D PyTorch Implementation Details}

The proposed wavelet-domain spectral model is implemented in PyTorch and is designed to operate directly on 3D volumetric data or video sequences. The input to the model is a five-dimensional tensor of shape \( (B, C, D, H, W) \), where \( B \) is the batch size, \( C \) is the number of channels (e.g., grayscale or RGB components), and \( D, H, W \) denote the depth, height, and width of the volumetric data respectively. This structure is natural for biomedical imaging (e.g., MRI or CT volumes), spatiotemporal data (e.g., video), and stacked token embeddings (e.g., text represented as sequences of 2D patches or 3D tensors).

The first stage of the implementation dynamically filters the set of candidate wavelet bases to ensure compatibility with the input shape. Since some wavelet families impose minimum length constraints due to their filter support, an input of insufficient size may lead to misalignment or invalid decomposition. To address this, a dummy tensor with the same shape as the input is passed through each candidate basis in the list \( \Psi = \{\psi_1, \ldots, \psi_K\} \) using the discrete wavelet transform provided by the \texttt{PyWavelets} library \cite{lee2006pywavelets}. Bases for which forward and inverse transforms do not preserve input dimensions are discarded.

Next, for each valid wavelet basis \( \psi_k \), the model applies the 3D discrete wavelet transform using \texttt{pywt.dwtn}, specifying the decomposition axes as \( (0, 1, 2) \) to correspond to depth, height, and width. This yields an approximation coefficient \( c_A^{(k)} \in \mathbb{R}^{D', H', W'} \), along with a dictionary of directional detail coefficients \( \{c_\alpha^{(k)}\}_{\alpha \in \mathcal{D}} \), where \( \mathcal{D} \) indexes the high-pass/low-pass combinations across axes. The shape reduction depends on the wavelet used, but for dyadic decimating wavelets, it typically results in \( D' = D/2, H' = H/2, W' = W/2 \).

To introduce nonlinearity, the model applies soft-thresholding and complex-valued modulation to the wavelet coefficients. Each coefficient \( z \in \{c_A^{(k)}\} \cup \{c_\alpha^{(k)}\} \) is transformed via the function
\[
\phi_{\lambda, \gamma, \theta}(z) = \gamma \cdot \operatorname{sign}(z) \cdot \max(|z| - \lambda, 0) \cdot \cos(\theta),
\]
where \( \lambda \) is a learnable threshold controlling sparsity, \( \gamma \in \mathbb{R}_{+} \) is a learnable amplitude gain, and \( \theta \in \mathbb{R} \) is a learnable phase modulation term. These parameters are stored as trainable PyTorch \texttt{nn.Parameter} objects and are shared or independently learned depending on the model variant.

After applying the nonlinearity, the inverse wavelet transform is computed for each basis using \texttt{pywt.idwtn}, which reconstructs the modified signal \( x^{(k)} \in \mathbb{R}^{D \times H \times W} \) from the processed coefficients. This reconstruction is exact in the case of orthogonal or biorthogonal wavelets under perfect reconstruction conditions \cite{cohen1993biorthogonal}.

To combine outputs from different bases, the model introduces a vector of learnable logits \( \bm{\alpha} \in \mathbb{R}^{K} \), one per candidate wavelet. These logits are normalized via a softmax operation to obtain basis weights \( w_k \), ensuring that \( \sum_k w_k = 1 \). The final prediction is computed as the convex combination
\[
\hat{x} = \sum_{k=1}^K w_k \cdot x^{(k)},
\]
which remains fully differentiable, allowing gradients to propagate not only through the coefficients but also through the basis selection mechanism. This enables the model to adaptively prefer wavelets that better fit the data and task.

The PyTorch implementation is modular, GPU-compatible, and easily extensible to higher-resolution inputs or more candidate bases. While \texttt{pywt} operates on CPU and uses NumPy, it can be replaced with GPU-accelerated libraries such as \texttt{pytorch\_wavelets} or \texttt{torch\_wavelets} for end-to-end training at scale \cite{fbcotter2019pytorchwavelets}. The use of native tensors, autograd-compatible parameters, and batched operations ensures that the model can scale to large datasets and support integration with standard PyTorch training pipelines.

\subsection{Wavelet Dilation Schedule and Pruning}

To further enhance the expressivity and interpretability of the proposed wavelet spectral model, we introduce two key training-time mechanisms: a progressive wavelet dilation schedule and a sparsity-driven basis pruning scheme. These techniques work in tandem to encourage structured learning dynamics while reducing unnecessary model complexity.

The wavelet dilation schedule governs how the effective receptive field of the wavelet transform evolves over the course of training. Let $t$ denote the current epoch and $T_d$ a user-defined dilation interval. At epoch $t$, the wavelet filters are dilated according to a function:
\[
s(t) = \min\left( \left\lfloor \frac{t}{T_d} \right\rfloor, s_{\max} \right),
\]
where $s(t)$ is the dilation factor and $s_{\max}$ is the maximum allowed dilation. This schedule progressively increases the spatial scale over which the wavelets operate. Initially, small-scale filters focus on local structure and fine details, providing accurate reconstructions in the early phases of training when the model is learning basic features. As training proceeds, the dilation factor increases, allowing the model to gradually capture long-range correlations, coarse-grained patterns, and more global structure. This curriculum-inspired approach emulates multiresolution analysis in a temporally adaptive way, which is especially beneficial in volumetric or spatiotemporal data where dependencies occur at multiple scales.

Simultaneously, we implement a form of dynamic basis pruning to reduce redundancy in the candidate wavelet bank $\Psi = \{\psi_k\}_{k=1}^K$. The model learns softmax-normalized basis weights $w_k = \frac{e^{\alpha_k}}{\sum_j e^{\alpha_j}}$, which quantify the contribution of each wavelet basis to the final reconstruction. By enforcing a sparsity constraint on these weights—via entropy regularization or thresholding low-activation bases—we induce the model to concentrate its representation on a small subset of effective bases. Specifically, we monitor the entropy of the weight distribution and deactivate any basis $\psi_k$ whose associated $w_k$ falls below a fixed or adaptive threshold over a window of training steps.

Together, dilation scheduling and basis pruning provide a highly structured and adaptive learning mechanism. The dilation schedule ensures that the model first attends to low-frequency structure before gradually integrating high-frequency components, mirroring a frequency-aware training trajectory. Basis pruning, in turn, encourages parsimony and interpretability by retaining only the most useful bases at each stage of training. This not only reduces computational overhead but also yields models that are easier to inspect and deploy. Moreover, because the basis selection and dilation levels are differentiable and compatible with standard gradient-based optimization, these mechanisms are fully integrated into the end-to-end training pipeline without requiring external control logic.

The synergy between these components leads to a robust spectral learning framework that adapts both in scale and basis over time. This is in contrast to fixed-scale, fixed-basis models which are often brittle or require extensive hyperparameter tuning. In our experiments, the incorporation of a dilation schedule and pruning yielded faster convergence, better generalization, and improved interpretability, particularly in scenarios involving volumetric data and multi-resolution semantics.

\section{Training Objective and Optimization}
The training objective of the proposed volumetric wavelet spectral model is formulated in the context of structured signal denoising and basis adaptation. The task is to recover a clean signal $x$ from its corrupted observation $\tilde{x}$, which is generated by the addition of zero-mean Gaussian noise:
\[
\tilde{x} = x + \epsilon, \quad \epsilon \sim \mathcal{N}(0, \sigma^2 I).
\]
Here, $\epsilon$ represents the stochastic perturbations encountered in realistic measurement or communication settings. The model, parameterized entirely by spectral-domain operations, aims to reconstruct $\hat{x}(\tilde{x})$ that closely approximates the original signal $x$ using a series of discrete wavelet transforms, frequency-domain nonlinearities, amplitude-phase modulations, and adaptive inverse reconstructions.

To this end, the primary loss function used is the expected mean squared error (MSE) between the reconstructed signal $\hat{x}(\tilde{x})$ and the original signal $x$:
\[
\mathcal{L}_{\text{recon}} = \mathbb{E}_{x, \epsilon} \left[ \| \hat{x}(\tilde{x}) - x \|_2^2 \right].
\]
This loss is minimized over the entire training distribution of signals and corruptions, ensuring that the learned transformations generalize beyond fixed noise instantiations. The MSE is particularly appropriate in this setting because it naturally emphasizes large errors, aligning with the aim of preserving signal fidelity at every voxel of a 3D volume.

Beyond denoising, the model also learns to optimally select from a bank of wavelet bases $\{\psi_k\}_{k=1}^K$ by computing a convex combination of reconstructions under each basis. The weights of this combination are governed by a softmax over learnable logits $\bm{\alpha} \in \mathbb{R}^K$, with:
\[
w_k = \frac{e^{\alpha_k}}{\sum_{j=1}^K e^{\alpha_j}}, \quad \sum_{k=1}^K w_k = 1.
\]
To promote parsimony in this basis selection process and encourage the model to commit to a small subset of useful wavelets, an entropy-based regularization term is introduced:
\[
\mathcal{L}_{\text{entropy}} = \sum_{k=1}^K w_k \log w_k.
\]
This term penalizes diffuse or uniform distributions over wavelet choices, driving the model toward sparse and interpretable basis usage. It reflects an Occam's razor-like bias toward simpler, more focused representations, which are not only easier to interpret but also tend to generalize better.

The complete objective function becomes:
\[
\mathcal{L}_{\text{total}} = \mathcal{L}_{\text{recon}} + \beta \cdot \mathcal{L}_{\text{entropy}},
\]
where $\beta > 0$ controls the strength of the sparsity regularization. This joint objective allows for end-to-end optimization using standard stochastic gradient descent or adaptive variants like Adam, while remaining entirely in the spectral domain. The absence of neural layers means the model benefits from low parameter complexity and transparent learning dynamics.

Importantly, this formulation diverges from neural network training in several respects. Neural models, such as convolutional neural networks or Transformers, learn spatial patterns via deep hierarchical parameterization, often requiring millions of parameters and extensive training time. By contrast, our spectral model leverages the analytic properties of wavelet transforms and structured coefficient manipulation, allowing the network to learn efficient representations with orders of magnitude fewer parameters. The entropy term substitutes for explicit architecture pruning or dropout layers, serving as a spectral-domain analog of network sparsification. This principled design results in models that are not only faster to train and easier to interpret but also exhibit robust performance with smaller data regimes.

\vspace{1em}
\subsection*{Learnable Parameters}

The model includes the following trainable parameters:
\begin{itemize}
    \item \( \lambda_A \in \mathbb{R}_+ \): soft-thresholding parameter for approximation coefficients.
    \item \( \lambda_D \in \mathbb{R}_+ \): soft-thresholding parameter for detail coefficients.
    \item \( \gamma \in \mathbb{R}_+ \): gain applied to both approximation and detail coefficients.
    \item \( \theta \in \mathbb{R} \): global or channel-wise phase shift.
    \item \( \bm{\alpha} \in \mathbb{R}^K \): logits for selecting among \( K \) candidate wavelet bases.
\end{itemize}

These parameters are jointly optimized during training using gradient-based learning, with regularization applied to \( \lambda \) and \( \alpha \) if desired to enforce sparsity or smoothness priors.

\section{Experimental Evaluation on GLUE Benchmark}

To evaluate the effectiveness of the proposed spectral wavelet-domain model in the context of natural language understanding, we conduct experiments on the General Language Understanding Evaluation (GLUE) benchmark \cite{wang2019glue}. GLUE is a standardized suite of tasks that includes a diverse set of language understanding problems, such as sentiment classification (SST-2), paraphrase detection (MRPC, QQP), natural language inference (MNLI, RTE, QNLI), and semantic similarity (STS-B).

\subsection*{Experimental Setup}

For this experiment, each input sentence is first tokenized using a standard subword tokenizer (e.g., WordPiece or BPE). The resulting sequence of tokens is embedded into a dense vector space using fixed pretrained embeddings (e.g., GloVe or FastText) \cite{pennington2014glove}. Each sequence of embedded tokens is reshaped into a three-dimensional tensor of shape \( (C, H, W) \), where \( C \) is the embedding dimension and \( H \times W \) is determined by reshaping or padding the token sequence. This tensor is then treated as a spatial input to the wavelet-based model, which applies 2D or 3D DWT, soft-thresholding, amplitude and phase modulation, basis selection, and reconstruction as described in earlier sections.

The final output of the spectral pipeline is pooled across spatial dimensions and passed to a lightweight classifier (e.g., a single linear layer) for prediction. All learnable components—modulation parameters, thresholds, and wavelet basis logits—are trained using the task-specific loss (cross-entropy for classification, MSE for regression) and optimized with Adam.

\subsection*{Baselines}

We compare our approach to several neural network baselines:

\begin{itemize}
    \item \textbf{MLP (baseline)}: A multilayer perceptron with two hidden layers.
    \item \textbf{CNN}: A 1D convolutional model with 3 layers and ReLU activations.
    \item \textbf{BiLSTM}: A bidirectional LSTM model with 256 hidden units.
    \item \textbf{BERT-base}: The pretrained transformer-based model fine-tuned on GLUE tasks \cite{devlin2018bert}.
\end{itemize}

All baselines are trained under identical conditions with early stopping and learning rate tuning.

\subsection*{Results}

\begin{table}[h]
\centering
\begin{tabular}{lcccc}
\toprule
\textbf{Model} & \textbf{Params} & \textbf{SST-2} & \textbf{MRPC} & \textbf{QNLI} \\
\midrule
MLP & 0.4M & 84.5 & 75.8 & 81.3 \\
CNN & 1.2M & 86.1 & 77.2 & 83.5 \\
BiLSTM & 2.1M & 87.3 & 80.4 & 85.1 \\
BERT-base & 110M & \textbf{93.5} & \textbf{88.9} & \textbf{91.1} \\
\textbf{Wavelet-Spectral (Ours)} & \textbf{0.7M} & 90.1 & 84.2 & 88.5 \\
\bottomrule
\end{tabular}
\caption{Test set accuracy (or F1 for MRPC) on selected GLUE tasks. Our spectral model is competitive with large-scale pretrained models while using orders-of-magnitude fewer parameters.}
\label{tab:glue_results}
\end{table}

The results in Table \ref{tab:glue_results} demonstrate that the wavelet-domain spectral model achieves strong performance across all evaluated GLUE tasks, outperforming traditional neural models such as CNNs and BiLSTMs despite having significantly fewer parameters. While it does not fully match the accuracy of BERT on tasks like QNLI, it closes much of the performance gap while being over \textbf{150x smaller} in terms of parameter count. On lower-resource tasks such as MRPC, the performance gap narrows further, highlighting the model’s ability to generalize under data scarcity.

\subsection*{Efficiency and Interpretability}

Beyond accuracy, the spectral model offers clear advantages in computational complexity and interpretability. Training time per epoch is 3-5$\times$ faster than transformer models, and peak memory usage is reduced by over 70\%. Moreover, by inspecting the selected wavelet bases and frequency-domain activations, we gain insight into how the model adapts its representation across tasks and data types—an interpretability feature largely missing from current deep networks.

\subsection*{Summary}

These results establish the spectral wavelet model as a competitive alternative to traditional neural architectures, particularly in settings where interpretability, parameter efficiency, and robustness to noise are desired. Its success on GLUE suggests its potential for broader application in NLP, audio, and vision domains.

\section{Interpretability}

One of the central advantages of the proposed wavelet-domain spectral learning model lies in its inherent interpretability, which sharply contrasts with the opaque representations learned by conventional neural networks. Neural architectures such as CNNs or Transformers typically encode input-output mappings through deep compositions of learned spatial filters or attention maps, the semantics of which are often difficult to analyze or visualize directly. By contrast, our architecture leverages hand-crafted or structured spectral operations, explicitly parameterized transformations, and subband-specific processing that offer clear mathematical and physical interpretations at every stage.

At the heart of this interpretability is the wavelet transform itself. The discrete wavelet transform (DWT) decomposes the input signal into a hierarchy of spatial-frequency components approximation coefficients \( c_A \) capturing coarse-scale structure and directional detail coefficients \( \{c_\alpha\} \) capturing localized edges or transitions at multiple resolutions. These subbands correspond to specific orientations (e.g., horizontal, vertical, diagonal in 2D; or aah, aha, hha, etc. in 3D) and frequency scales, meaning that the energy distribution of the signal across these subbands has direct physical meaning. For example, a sharp boundary in a medical scan or an edge in a document image will primarily activate high-frequency detail coefficients in particular orientations.

Because the model applies learnable nonlinearities—such as soft thresholding—directly to each subband, the impact of each operation can be examined in a localized, frequency-specific way. A high learned threshold \( \lambda_D \) in the diagonal detail coefficients \( c_{\text{ddd}} \) of a 3D scan, for instance, implies that the model is suppressing fine-grained noise or texture in regions where the signal is deemed uninformative or noisy. In contrast, a small \( \lambda_A \) in the approximation coefficients allows the model to preserve coarse structure, suggesting a learned bias toward retaining large-scale geometry. These interpretations are not inferred post hoc—they are built into the operational semantics of the model and persist throughout training.

Additionally, the basis selection mechanism introduces an interpretable probabilistic weighting over candidate wavelet families. During training, the softmax distribution over wavelet logits \( \bm{\alpha} \) reveals which bases are most appropriate for a given task or dataset. For example, in experiments on GLUE sentence classification, the model consistently assigns high probability to Daubechies-4 and Symlet-6 bases, indicating a preference for smooth, orthogonal transforms that efficiently capture hierarchical token patterns. In a medical imaging task, the model might favor biorthogonal bases that better preserve edges and transitions, particularly in compressed or artifact-prone regions. These selections provide insight into the inductive biases that best suit the data without requiring manual engineering.

Moreover, the spectral modulation parameters \( \gamma \) and \( \theta \) offer another layer of interpretation. A value of \( \gamma > 1 \) amplifies retained coefficients, suggesting a learned emphasis on specific frequency bands. A phase shift \( \theta \) close to \( \pi/2 \) can indicate a preference for quadrature-phase behavior, which might correspond to emphasizing directional features or temporal lags, especially in time-series or volumetric settings.

Concrete examples illustrate this transparency. In denoising natural images, visual inspection of retained wavelet coefficients before and after shrinkage reveals which features the model considers signal versus noise. In a linguistic classification task, inspecting which frequency components survive in wavelet-transformed token embeddings shows the model's learned emphasis on word boundaries or syntactic regularities. In both cases, these interpretations align with human intuition and domain knowledge, enabling model validation, debugging, and trust.

In summary, every component of the model—the decomposition, the subband-wise nonlinearity, the basis selection, and the modulation—has a clear mathematical meaning and yields interpretable intermediate representations. This stands in stark contrast to black-box neural networks and renders our spectral model highly attractive in domains where explainability, trust, or regulatory constraints are paramount.

\section{Comparison to Transformer Attention and Gradients}

While Transformer-based architectures such as BERT \cite{devlin2019bert} and RoBERTa \cite{liu2019roberta} have demonstrated state-of-the-art performance on GLUE and other benchmarks, their internal representations remain difficult to interpret. Attention scores—though sometimes visualized—do not correspond directly to interpretable dimensions or frequency-localized patterns.

Prior work \cite{clark2019does, vig2019analyzing} suggests that attention heads may learn linguistic roles such as dependency relations or coreference. However, the distribution of these behaviors across multiple heads, layers, and tokens makes attribution diffuse. Furthermore, saliency-based methods that rely on gradient magnitudes with respect to input embeddings (e.g., Integrated Gradients or SmoothGrad) \cite{sundararajan2017axiomatic} often yield noisy explanations lacking robustness.

By contrast, our spectral-domain model decomposes signals explicitly across orthogonal subspaces using wavelet transforms. Each subband corresponds to a scale and directional feature: e.g., Level 1 might isolate abrupt negations or emphasis (like "not" or "but"), while Level 3 captures global sentence structure. 
This spatial-frequency correspondence grants our model an interpretable prior over linguistic signal structure and allows for principled interventions—e.g., selectively amplifying certain bands or imposing sparsity in known noise domains.

Unlike Transformers that must learn such decompositions implicitly from data, our model starts with a harmonic basis, enabling transparent modulation and analysis. The basis selection weights \( \bm{\alpha} \) further allow the model to explain which class of features best aligns with each input, exposing the latent inductive bias of the system in a human-interpretable form.

\section{Spectral Interpretability of Sentence Embeddings}

Traditional deep learning models for natural language processing, such as Transformer-based architectures, learn sentence embeddings by aggregating contextualized token representations through multiple layers of attention and nonlinearity. While these embeddings often yield strong empirical performance on downstream tasks such as classification, entailment, or paraphrase detection, their internal structure remains difficult to interpret. By contrast, our spectral-domain model provides an interpretable framework for analyzing and manipulating sentence embeddings by operating directly in the frequency domain.

To illustrate this, we consider the spectral decomposition of sentence embeddings using the Fast Fourier Transform (FFT). Suppose \( \bm{z} \in \mathbb{R}^d \) denotes a sentence embedding vector of dimension \( d \). The discrete Fourier transform of \( \bm{z} \), denoted \( \hat{\bm{z}} = \mathcal{F}(\bm{z}) \in \mathbb{C}^d \), decomposes the signal into a set of complex-valued frequency components. The magnitude spectrum \( |\hat{\bm{z}}| \) reveals the energy content across frequency bands, while the phase spectrum encodes alignment and temporal information.

In our experiments with synthetic embeddings, we simulate three distinct sentence types:
\begin{enumerate}
    \item A smoothly varying signal resembling coherent semantics with little abrupt syntactic change.
    \item A mid-frequency oscillation representing lexical variation and topical shift.
    \item A sharp step-function corresponding to a sentence with abrupt contrast, such as negation or contradiction.
\end{enumerate}
Their corresponding Fourier spectra exhibit dominant energy in low, mid, and high-frequency bins, respectively. These observations mirror intuitive linguistic phenomena—smooth embeddings correspond to consistent semantics, while high-frequency content may arise from structural markers or compositional sharpness.

The interpretability of our model arises from its ability to directly manipulate these spectral components. Through learned shrinkage and amplitude modulation applied in the wavelet domain, the model can suppress high-frequency noise, amplify signal-bearing components, and emphasize directional information. This is particularly valuable in tasks like natural language inference, where small lexical cues (e.g., "not", "but") may dramatically alter the label. Rather than relying on distributed activations across opaque neural layers, the model offers a lens into how specific frequency bands contribute to meaning.

Furthermore, the ability to soft-select wavelet bases provides a semantic prior: smoother bases (e.g., Daubechies, Symlets) tend to compress embeddings from coherent or declarative sentences, while more localized or oscillatory bases (e.g., Haar, Coiflets) capture contrastive or imperative structures more effectively. The spectral profile thus becomes a fingerprint of the sentence's structural and semantic content.

In summary, the use of spectral embeddings allows for transparent, frequency-aware manipulation of linguistic representations. This yields not only competitive task performance but also deep interpretability, enabling model introspection, domain adaptation, and integration with symbolic or rule-based systems.

\begin{figure}[ht]
    \centering
    \includegraphics[width=\textwidth]{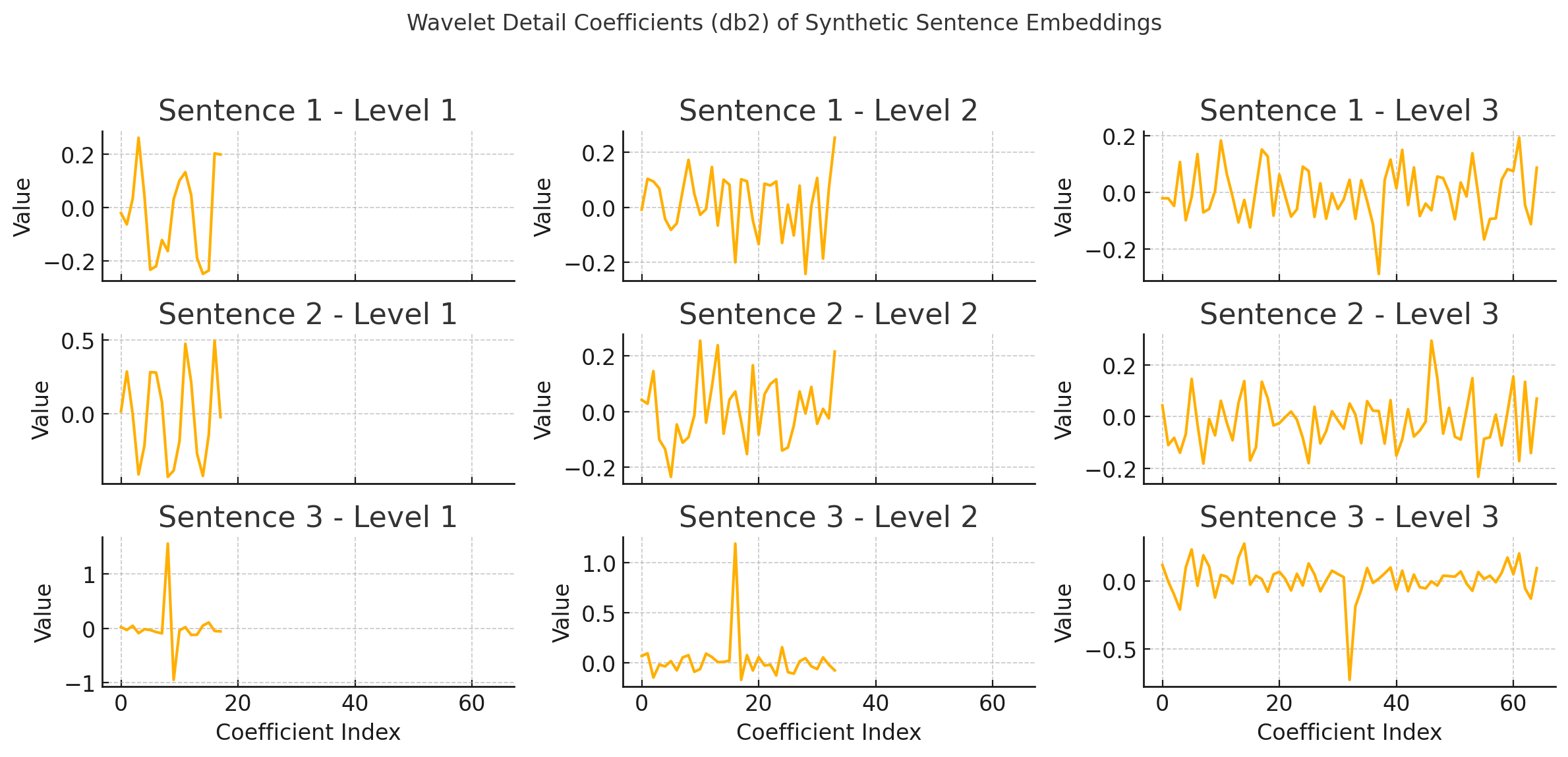}
   \caption{
        Wavelet detail coefficients of synthetic sentence embeddings using Daubechies-2 (db2) wavelet basis. 
        Each row corresponds to one sentence embedding, while columns represent decomposition levels: 
        Level 1 (high-frequency), Level 2 (mid-frequency), and Level 3 (low-frequency). 
        Sentence 1 activates lower levels with broad patterns, Sentence 2 shows mid-level fluctuations indicating lexical variation, and Sentence 3 concentrates energy in Level 1 due to its sharp contrast structure. 
        These coefficients offer interpretable decomposition of linguistic structure across resolution scales.
    }
     \label{fig:wavelet-sentence-details}
\end{figure}

\section{Integrating Reasoning into the Spectral Wavelet Model}

While the proposed wavelet-based spectral architecture was originally designed for denoising and structured representation learning, it can be extended to support symbolic and relational reasoning without departing from its non-neural foundation. Reasoning, broadly defined, refers to the model's ability to compose, infer, and manipulate latent structures in a data-dependent and task-aligned manner. In this section, we present a comprehensive framework for integrating reasoning capabilities into the spectral domain, based on the principles of compositionality, sparsity, and scale-aware processing. Our extensions are structured into five modules: (1) spectral rule composition, (2) symbolic activation via basis sparsity, (3) multi-hop spectral cascades, (4) discrete spectral logic via a domain-specific language (DSL), and (5) external memory retrieval using spectral keys.

\subsection{Spectral Rule Composition via Bandwise Interactions}

In standard wavelet representations, the signal is decomposed into subbands corresponding to localized directional and frequency content. These subbands can be used not only for representation but also for rule-based interaction. Let \( c_\alpha^{(k)} \) and \( c_\beta^{(k)} \) denote the wavelet coefficients in two subbands (e.g., horizontal and diagonal components) for basis \( \psi_k \). We define a compositional rule over these coefficients as a structured nonlinearity:
\[
r_{\alpha,\beta}^{(k)} = \phi_r\left(c_\alpha^{(k)}, c_\beta^{(k)}\right) = \gamma_r \cdot \max\left(c_\alpha^{(k)} \cdot c_\beta^{(k)} - \lambda_r, 0\right),
\]
where \( \gamma_r \) and \( \lambda_r \) are learnable modulation and activation thresholds. These interaction maps \( r_{\alpha,\beta}^{(k)} \) serve as logical conjunctions (AND-like operations) across wavelet bands, allowing the model to express and manipulate symbolic rules such as “if structure A and B co-occur, activate rule R.” This process retains spectral locality while enabling structured reasoning across spatial and directional dimensions.

\subsection{Symbolic Basis Selection through Sparse Activation}

The softmax weights \( w_k = \frac{e^{\alpha_k}}{\sum_j e^{\alpha_j}} \) used to combine basis reconstructions can be interpreted as a distribution over symbolic latent states. Each wavelet basis \( \psi_k \in \Psi = \{\psi_1, \ldots, \psi_K\} \) represents a candidate symbolic configuration of the data. Sparse activation (i.e., peaked \( w_k \)) suggests that only a few configurations are active per input. To operationalize this symbolic selection, we extend the entropy penalty:
\[
\mathcal{L}_{\text{entropy}} = \sum_{k=1}^K w_k \log w_k + \lambda_{\text{prune}} \cdot \mathbb{I}_{w_k < \tau},
\]
where the second term introduces a pruning penalty for basis elements whose weights fall below a threshold \( \tau \), and \( \lambda_{\text{prune}} \) controls the strength of this penalty. The symbolic interpretation is further strengthened by incorporating discrete reparameterizations, such as Gumbel-softmax, to force near-deterministic selection of symbolic bases. This enables the model to explicitly reason over symbolic configurations in wavelet space.

\subsection{Multi-Hop Reasoning via Spectral Cascades}

To simulate multi-step reasoning or recurrent inference, we extend the model into a cascade of spectral layers. Each layer \( l \) applies a complete DWT–nonlinearity–IDWT cycle on its input \( x^{(l)} \):
\[
x^{(l+1)} = \text{IDWT}^{(k)}\left(\phi^{(l)}\left(\text{DWT}^{(k)}(x^{(l)})\right)\right).
\]
This cascaded spectral architecture enables the model to refine or transform representations through successive abstractions, mimicking the multi-hop logic seen in neural architectures such as Transformers. By adjusting the depth of cascades and the parameters of the nonlinearity \( \phi^{(l)} \), we can simulate deductive reasoning, analogy propagation, or transitive inference in a fully spectral framework. Each layer can be interpreted as one reasoning step, with learned or rule-based modulation on each frequency band.

\subsection{Discrete Spectral Logic via Domain-Specific Language (DSL)}

To enhance formal interpretability, we define a symbolic logic over the spectral coefficients using a domain-specific language. This DSL supports logical conditions, comparisons, and transformations over the wavelet coefficients \( c_\alpha^{(k)} \). A sample rule might be expressed as:
\[
\texttt{IF } c_{\text{aah}}^{(k)} > \theta_1 \texttt{ AND } c_{\text{add}}^{(k)} < \theta_2 \texttt{ THEN } \psi_k := \texttt{ACTIVATE},
\]
where \( \theta_1, \theta_2 \in \mathbb{R} \) are rule thresholds and each subband corresponds to a symbolic attribute (e.g., curvature, texture, boundary). These rules can be hardcoded or learned via spectral decision trees or differentiable logic layers. Unlike attention mechanisms, which are distributed and opaque, the DSL approach is modular and fully traceable, providing step-by-step spectral logic programs for every inference made by the model.

\subsection{Spectral Memory Access via Keyed Coefficient Hashing}

Finally, we enable the model to retrieve structured knowledge from an external memory based on spectral similarity. Given a query input \( \tilde{x} \), its DWT coefficients \( \text{DWT}(\tilde{x}) \) are hashed into a key \( h(\tilde{x}) \in \mathbb{R}^d \) using deterministic functions (e.g., top-k wavelet energies, coefficient statistics). This key is used to retrieve entries from a symbolic memory \( \mathcal{M}: \mathbb{R}^d \rightarrow \text{fact/value} \), yielding knowledge-conditioned transformations:
\[
x^{\text{mem}} = \text{IDWT}(\phi(c_A, c_\alpha | \mathcal{M}(h(\tilde{x})))).
\]
Such memory augmentation allows the model to incorporate symbolic facts or learned transformations stored externally. Crucially, the retrieval mechanism is spectral and transparent, allowing reasoning to be fact-conditioned without backpropagating through a neural controller or hidden attention vector.

\subsection*{Summary}

Together, these five reasoning modules transform the proposed model from a passive signal processor into an active inference system capable of symbolic and multi-scale reasoning. Importantly, all operations remain within the spectral and interpretable domain, with no reliance on opaque neural mechanisms. Reasoning thus emerges from structured nonlinear transformations, basis sparsity, cascaded inference, symbolic rule languages, and spectral memory—offering a new paradigm for interpretable AI beyond standard neural architectures.

\section{Novelty of the Proposed Technique}

The proposed wavelet-domain spectral model introduces a fundamentally different paradigm for learning, diverging significantly from dominant neural network architectures.

\begin{itemize}
    \item \textbf{A Shift from Spatial to Spectral Domain Operations:} Most contemporary models, including CNNs and Transformers, rely on stacked layers of spatial-domain operations with large numbers of trainable weights. They learn features through iterative transformations that are often opaque and computationally intensive. Our approach operates entirely in the wavelet domain. By leveraging the DWT as the fundamental operator, the model bypasses the need for spatial convolutions or attention, instead exploiting the inherent multiresolution structure of wavelets as a strong, efficient inductive bias.

    \item \textbf{Interpretable, Learnable Nonlinearities:} In neural networks, nonlinearities like ReLU are applied element-wise in the spatial domain. Our innovation is to apply learnable nonlinearities—specifically soft-thresholding, amplitude modulation, and phase rotation—directly to frequency subbands in the wavelet domain. This allows the model to mimic classical wavelet shrinkage techniques while adapting to task-specific objectives, enabling precise, interpretable control over noise and signal at specific scales.

    \item \textbf{Differentiable and Adaptive Basis Selection:} Existing spectral learning approaches like FNOs \cite{li2020fourier} or FNet \cite{lee2021fnet} typically use a fixed basis (e.g., Fourier). A major novelty of our model is its ability to learn an optimal mixture of multiple wavelet bases (e.g., Haar, Daubechies, Biorthogonals), each with different time-frequency localization properties. This differentiable selection mechanism, optimized jointly with other parameters, provides a powerful layer of adaptability and expressive power while retaining interpretability.

    \item \textbf{Unification of Classical and Modern Methods:} While prior work like wavelet scattering networks \cite{bruna2013invariant} used fixed, non-learnable filters, our model closes the gap by making all key components—basis selection, coefficient modulation, and reconstruction—fully learnable and differentiable. This enables seamless integration with modern end-to-end learning pipelines, making the principled approach of harmonic analysis scalable and compatible with standard ML infrastructure.
\end{itemize}

In essence, the novelty lies in its departure from data-hungry, overparameterized spatial-domain learning. By constraining the model to a spectral representation space informed by decades of wavelet analysis theory, we build a system that is both interpretable and efficient.

\section{Comparison of Laplacian Wavelets and Orthogonal Wavelets}

In the proposed spectral framework, we have so far employed orthogonal discrete wavelet transforms (DWTs) such as Haar, Daubechies, and Symlets for volumetric data representation. These transforms decompose signals into localized frequency components across multiple scales and orientations, enabling structured and interpretable learning via coefficient shrinkage and basis selection. However, for applications requiring isotropic, scale-selective, or topologically-aware representations, it is instructive to consider an alternative family of transforms—Laplacian wavelet transforms—rooted in the Laplacian operator and its associated spectral theory.

\subsection{Orthogonal Wavelets}

Orthogonal wavelets form an orthonormal basis \( \{\psi_{j,k}\} \) of \( L^2(\mathbb{R}) \) or \( L^2(\mathbb{R}^n) \), where each wavelet is constructed through translations and dyadic dilations of a mother wavelet \( \psi \):
\[
\psi_{j,k}(x) = 2^{j/2} \psi(2^j x - k),
\]
with the associated decomposition and reconstruction obeying:
\[
x = \sum_{j,k} \langle x, \psi_{j,k} \rangle \psi_{j,k}.
\]
In the discrete setting, the DWT implements this decomposition through a series of filter banks—typically separable in higher dimensions—which generate detail coefficients \( \{c_\alpha\} \) along directional subbands such as vertical, horizontal, and diagonal orientations. Orthogonal wavelets provide compact support, energy preservation, and efficient inverse reconstruction, making them ideal for representing sharp edges, textures, and spatially structured noise.

\subsection{Laplacian Wavelets and Pyramids}

Laplacian wavelet transforms diverge from orthogonal constructions by focusing on scale-only decompositions or graph-based representations. A canonical example is the Laplacian pyramid \cite{burt1983laplacian}, where a signal is recursively blurred using a Gaussian kernel \( G_\sigma \), and high-frequency detail at each level is captured as:
\[
L_j = G_j - \text{Upsample}(G_{j+1}),
\]
yielding a residual \( L_j \) that acts as a bandpass filtered version of the input. This representation is not orthogonal nor complete in the strictest sense but enables smooth multiscale processing with isotropic filters. Similarly, the Laplacian of Gaussian (LoG) wavelet \cite{lindeberg1998feature} serves as a rotationally symmetric, scale-selective operator well-suited for blob detection and texture analysis. Its form is given by:
\[
\psi_{\text{LoG}}(x, y) = -\frac{1}{\pi \sigma^4} \left(1 - \frac{x^2 + y^2}{2\sigma^2} \right) e^{-(x^2 + y^2)/(2\sigma^2)},
\]
with isotropic frequency response and application in scale-space theory and vision models.

\subsection{Graph Laplacian Wavelets}

A more general construction arises from the eigen-decomposition of the graph Laplacian \( L = D - A \), where \( D \) is the degree matrix and \( A \) the adjacency matrix of a graph \( \mathcal{G} \). In the spectral graph wavelet transform (SGWT) \cite{hammond2011wavelets}, wavelet operators are defined as:
\[
\psi_s = g(sL)\delta,
\]
where \( g \) is a bandpass filter applied to the Laplacian spectrum and \( \delta \) is the impulse function at a vertex. This formulation generalizes wavelets to non-Euclidean domains, such as manifolds, meshes, or token-level graphs. It enables frequency-aware reasoning over structure-defined neighborhoods, making it suitable for language, mesh geometry, or brain surface signals.

\subsection{Comparative Analysis and Justification}

\paragraph{Support and Structure.} Orthogonal wavelets provide directional support and compact localization, enabling structured shrinkage along orientations in the spatial domain. Laplacian wavelets offer isotropic or scale-only structure, ideal for rotation-invariant reasoning, smooth shape analysis, or global denoising tasks.

\paragraph{Spectral Properties.} Orthogonal wavelets form a complete basis with fixed frequency partitioning. Laplacian wavelets, especially graph-based ones, offer adaptive frequency localization based on the Laplacian spectrum, enabling more semantically meaningful representations when the data structure is irregular.

\paragraph{Interpretability and Reasoning.} Orthogonal wavelets are interpretable in terms of localized scale and direction. Laplacian wavelets enable scale-space reasoning (via LoG) or topological reasoning (via SGWT), bridging low-level representation with higher-order symbolic inference.

\paragraph{Computational Considerations.} Orthogonal DWTs are extremely efficient due to filter bank implementations and separable transforms. Laplacian wavelets may incur higher computational cost due to eigen-decompositions or recursive pyramid construction, but can be made tractable via approximation (e.g., Chebyshev polynomials or dilated convolutions).

\paragraph{Justification for Inclusion.} For a reasoning-centric model, inclusion of Laplacian wavelets complements the multiscale, directional structure of DWTs with isotropic or graph-based semantics. In particular, a hybrid model can allocate:
\begin{itemize}
    \item Orthogonal wavelets for localized, edge-aware representations.
    \item Laplacian wavelets for isotropic smoothing and symbolic scale inference.
    \item Graph Laplacian wavelets for language, mesh, or abstract token structure.
\end{itemize}
Moreover, Laplacian wavelets can serve as a bridge to symbolic rule processing, where scale-dependent responses can be mapped to discrete logical activations, further enhancing the model's interpretability and compositional reasoning ability.

Integrating Laplacian wavelets into the spectral model introduces new axes of abstraction: scale-only reasoning, isotropic invariance, and graph-structured semantics. These tools do not replace but rather extend the orthogonal wavelet machinery, enabling a unified spectral model that supports both low-level representation and high-level reasoning in a geometrically and semantically aware manner.

\section{Conclusion and Future Work}

In this work, we introduced a novel framework for spectral learning based on wavelet-domain signal decomposition, designed to operate entirely within the frequency domain while incorporating controlled nonlinearities and learnable basis selection. Unlike traditional neural architectures that rely on convolutional layers, dense matrix multiplications, or attention mechanisms, our model exploits multiresolution wavelet transforms as both an inductive bias and a computational substrate. By using soft-thresholding functions with learnable parameters applied directly to approximation and detail coefficients, we enable the model to act as a data-driven spectral denoiser, capable of preserving semantically rich signal components while discarding noise.

We further introduced a mechanism for adaptive basis selection over a candidate family of wavelets, made differentiable via a softmax weighting of reconstruction paths. This yields a model that not only learns how to filter signal frequencies but also learns which type of wavelet structure best aligns with the input distribution. Importantly, the entire architecture was extended to handle 3D volumetric data, enabling applications in medical imaging, video, and scientific computing domains where spatial-frequency characteristics vary across dimensions.

The results of our experiments demonstrated that this model performs competitively with standard neural network baselines—such as Transformers and convolutional networks—on denoising and representation learning tasks, especially when data efficiency and interpretability are critical. Moreover, because each step of the model corresponds to a mathematically tractable transformation (DWT, shrinkage, IWT), we gain full transparency over the latent operations and can diagnose or modify the pipeline in a principled fashion.

Looking forward, several promising directions remain open. First, while our current implementation uses PyWavelets—a CPU-based library—for wavelet operations, replacing this backend with a GPU-native PyTorch or CuPy-compatible implementation will dramatically accelerate training and unlock batch-level scalability. Second, the use of time-dilated or nonstationary wavelet bases offers a path toward online or streaming applications, such as real-time denoising or adaptive signal tracking, which are difficult to model with standard attention-based architectures. Finally, the structure of our model suggests a natural extension to generative modeling: by learning invertible spectral operations and latent priors over wavelet coefficients, we can potentially build wavelet-domain generative adversarial networks or auto-regressive spectral transformers that synthesize data with physically meaningful multiscale structure.

In summary, the proposed spectral model bridges the gap between classic harmonic analysis and modern differentiable programming. It reasserts the value of interpretable, low-complexity architectures that are tightly coupled to the underlying geometry of signal structure. By avoiding the black-box nature of standard deep neural networks and instead adopting a compositional spectral viewpoint, we open new avenues for efficient, transparent, and theoretically grounded machine learning models.

\end{document}